\documentclass{ecai}
\usepackage{times}
\usepackage{graphicx}
\usepackage{latexsym}

\usepackage{amsmath,amssymb}
\DeclareMathOperator{\E}{\mathbb{E}}

\begin{document}

\title{Playing Atari Games with Deep Reinforcement Learning and Human Checkpoint Replay}

\author{Ionel-Alexandru Hosu\institute{University Politehnica of Bucharest, Romania, email: ionel.hosu@stud.acs.upb.ro} \and Traian Rebedea\institute{University Politehnica of Bucharest, Romania, email: traian.rebedea@cs.pub.ro} }

\maketitle

\bibliographystyle{ecai}

\begin{abstract}
  This paper introduces a novel method for learning how to play the most difficult Atari 2600 games from the Arcade Learning Environment using deep reinforcement learning. The proposed method, called \textit{human checkpoint replay}, consists in using checkpoints sampled from human gameplay as starting points for the learning process. This is meant to compensate for the difficulties of current exploration strategies, such as $\varepsilon$-greedy, to find successful control policies in games with sparse rewards. Like other deep reinforcement learning architectures, our model uses a convolutional neural network that receives only raw pixel inputs to estimate the state value function. We tested our method on Montezuma's Revenge and Private Eye, two of the most challenging games from the Atari platform. The results we obtained show a substantial improvement compared to previous learning approaches, as well as over a random player. We also propose a method for training deep reinforcement learning agents using human gameplay experience, which we call human experience replay.
\end{abstract}

\section{INTRODUCTION}
General game playing is an extremely complex challenge, since building a model that is able to learn to play any game is a task that is closely related to achieving artificial general intelligence (AGI). Video games are an appropriate test bench for general purpose agents, since the wide variety of games allows solutions to use and hone many different skills like control, strategy, long term planning and so on. Designed to provide enough of a challenge to human players, Atari games in particular are a good testbed for incipient general intelligence. With the release of the Arcade Learning Environment (ALE) \cite{bellemare2012arcade} in 2012, general game playing started to gain more popularity. ALE is an emulator for the Atari 2600 gaming platform and it currently supports more than 50 Atari games. They are somewhat simple, but they provide high-dimensional sensory input through RGB images (game screen).

Deep learning models have achieved state-of-the-art results in domains such as vision \cite{krizhevsky2012imagenet, DBLP:journals/corr/SzegedyLJSRAEVR14, DBLP:journals/corr/HeZRS15} and speech recognition \cite{DBLP:journals/corr/AmodeiABCCCCCCD15}. This is due to the ability of models such as convolutional neural networks to learn high-level features from large datasets. Along with these achievements, reinforcement learning also gained a lot of ground recently. These powerful models helped reinforcement learning where it struggled the most, by providing a more flexible state representation. As a consequence, tasks such as learning multiple Atari games using a single, unmodified architecture became achievable through deep reinforcement learning \cite{mnih2013playing, mnih2015human}. 
However, in environments characterized by a sparse or delayed reward, reinforcement learning alone is still struggling. This is caused mostly by naive exploration strategies, such as $\varepsilon$-greedy \cite{sutton1998reinforcement}, that fail to find successful policies to discover an incipient set of rewards. This is the case for the most difficult video games from the Atari platform, such as Montezuma's Revenge and Private Eye, that prove to be too challenging for all current approaches.

In a different context, AlphaGo \cite{silver2016mastering}, a system combining reinforcement learning with Monte Carlo Tree Search, defeated Lee Sedol, one of the top Go players in the world. Nearly 20 years after Garry Kasparov was defeated by Deep Blue \cite{campbell2002deep}, this represented an important milestone in the quest for achieving artificial general intelligence. Together with the launch of the OpenAI Gym \cite{brockman2016openai}, it is one of the most significant advances that reinforcement learning made in the last couple of years.

This paper demonstrates that it is possible to successfully use a learning approach on the most complex Atari video games, by introducing a method called \textit{human checkpoint replay}. Our method consists of using checkpoints sampled from the gameplay of a human player as starting points for the training process of a convolutional neural network trained with deep reinforcement learning \cite{mnih2013playing, mnih2015human}.
% IH: Note to self: ^aici mai trebuie adaugat.

The paper proceeds as follows. In section 2, we present the most relevant results on learning for Atari video games. Section 3 follows with a more in-depth analysis and explanation of the current results for Atari games. We also motivate our choice of games for evaluating our architecture, providing a brief description of the difficulties encountered in these games. In section 4 we provide a thorough description of the proposed methods and deep reinforcement learning architecture. Section 5 presents the results of our approach, as well as a detailed discussion on the performed experiments.

\section{RELATED WORK}
After the release of the Arcade Learning Environment, there have been numerous approaches to general game playing for Atari games. Approaches such as SARSA and contingency awareness \cite{bellemare2012investigating} delivered promising results, but were far from human-level performance. The use of neuro-evolution \cite{hausknecht2012hyperneat, hausknecht2014neuroevolution} on the Atari platform dramatically improved these results, but playing Atari games as well as a human player still seemed unachievable.

The first method to achieve human-level performance in an Atari game is deep reinforcement learning \cite{mnih2013playing, mnih2015human}. It mainly consists of a convolutional neural network trained using Q-learning \cite{watkins1992q} with experience replay \cite{lin1993reinforcement}.
The neural network receives four consecutive game screens, and outputs Q-values for each possible action in the game. Experience replay is used in order to break the correlations between consecutive updates, as Q-learning would prove unstable in an online setting. The most important aspect of this approach is that it can be used to construct agents that do not possess any prior domain knowledge, thus rendering them capable of learning to perform multiple different tasks.

After this first success of deep reinforcement learning \cite{mnih2013playing, mnih2015human}, a number of improvements have been made to the original architecture. The fact that its convergence was slow and it took multiple days to train a neural network on a single game motivated the development of a distributed version of deep reinforcement learning \cite{nair2015massively} which reduces the training times and improves the existing results.

Another notable improvement came from the realization that the Q-learning algorithm sometimes performs poorly by overestimating action values \cite{hasselt2010double}. This issue may be solved by employing double Q-learning \cite{van2015deep} - using it together with deep reinforcement learning on the Atari domain fixed the overestimation problem that appeared in some of the games.
% TR: poate aici si la sectiunea anterioara ar trebui sa spunem cam ce imbunatatiri au fost aduse. Scoruri cu x% mai mari pentru 5-6 jocuri, pentru jumatate din jocuri, etc. Dar merge si asa, este doar o parere.

Another notable approach to Atari game playing is the bootstrapped deep Q-network (DQN) \cite{osband2016deep}, which proposes a novel and computationally efficient method of exploration. Its main contribution is to find an alternative for simple, inefficient exploration strategies, such as $\varepsilon$-greedy. To achieve this, bootstrapped DQN produces distributions over Q-values instead of Q-values. Sampling from these distributions allows the model to renounce using exploration strategies.
% TR: ultima fraza este neclara

The current state-of-the-art on Atari games is achieved using a method called prioritized experience replay \cite{schaul2015prioritized}. It starts from the assumption that not all the transitions present in the replay memory have the same importance. The agent learns more effectively from some transitions, other ones being redundant, not relevant, etc. The method proposes a prioritization regarding how often transitions are used for updates in the network based on the magnitude of their temporal difference (TD) error \cite{van2013planning}. Prioritized replay leads to an improvement in 41 out of the 49 games, delivering human-level performance in 35 of these games.

Although the aforementioned methods brought considerable improvements over the original deep reinforcement learning architecture, there still are some Atari games for which none of the previously published methods are able to learn whatsoever. These games feature a sparse reward space and are more complex in many aspects than the vast majority of games from the Atari platform. 
% TR: poate ar trebui sa atragem atentia undeva ca sunt mai asemanatoare cu realitatea sau cu situatiile cat de cat reale mai complexe, unde reward-ul nu este dens sau usor de atins
We propose a new method called \textit{human checkpoint replay} that, when used together with deep reinforcement learning, is able to learn successful policies for the most difficult games from the Atari platform, thus resulting in significantly improved performance compared to prior work.

\section{BACKGROUND}
Before proceeding to describe our approach, it is important to perform a more detailed analysis of the games used in our experiments; the discussion focuses on highlighting some of the aspects that make them so challenging. To evaluate our approach, we have chosen Montezuma's Revenge and Private Eye - two of the most difficult games from the Atari platform. Therefore our analysis focuses on these two games; however the main points are also valid for other challenging games where there is no known learning policy which achieves a better score than a random agent.

% TR: momentan nu vad poza referita in text, ar trebui sa facem asta
\begin{figure}
\centerline{\includegraphics[height=2.4in]{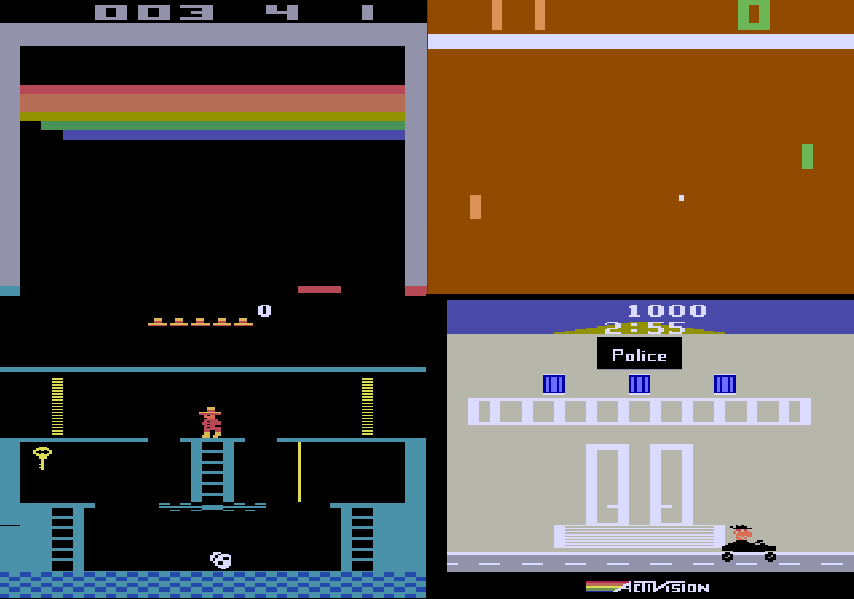}}
\caption{Screen shots from four Atari 2600 Games: (\textit{Left-to-right, top-to-bottom}) Breakout, Pong, Montezuma's Revenge, Private Eye} 
\label{procstructfig}
\end{figure}

\subsection{Montezuma's Revenge}
Montezuma's Revenge is a game from the Atari 2600 gaming console that features a human-like avatar moving across a series of 2D rooms that form labyrinth. In order to advance in the game, the player must move in a consistent manner, jump over obstacles, climb ladders, avoid or kill monsters, and collect keys, jewels and other artifacts which provide a positive reward by increasing the game score. Some of these collectible artifacts grant additional abilities - for example, collecting a key enables the player to open a door upon contact. However, after opening a door, the player loses the key and needs to collect additional keys for opening any other doors. Other collectible items include a torch which lights up dark rooms and swords which kill monsters. The game consists of three levels, each of them containing 24 rooms.

An important characteristic of the game is that collectible items are sparse and thus the reward space is also very sparse. When the game starts, in order to collect the first reward - represented by a key - the avatar is required to descend on two ladders, walk across the screen and over a suspended platform, jump over a monster and climb another ladder. After collecting the key, the avatar needs to return close to the starting point, where two doors are available which can now be opened using the collected key. The player starts the game with five "lives", and each time it loses one life, the avatar is respawned in the same room. For optimal play, a memory component is required, as the player does not possess information about other rooms in terms of rewards collected or monsters killed. The only information available on the screen apart from the environment is the game score, the number of remaining lives and the artifacts currently held by the avatar.

All current approaches using deep reinforcement learning fail to learn any successful control policies for Montezuma's Revenge. This happens mostly due to the $\varepsilon$-greedy strategy failing to explore the game in a consistent and efficient manner. Every four frames, a DQN agent has to choose between 18 different actions. Given how, in order to receive the first positive reward, the player is required to perform a complex and consistent sequence of actions, using such a simple exploration strategy makes learning virtually impossible. It is also worth mentioning that the way in which such a simple strategy explores an environment like Montezuma's Revenge does not resemble the way in which a human player does it. This is due to two factors. First, there is a strong correlation between multiple successive actions of a human player in the video game, even when the player does not know how to play the game yet. This is because the exploration exhibited by a human player is not random, but influenced by the current state of the environment. Second, a human player always makes use of commonsense knowledge when dealing with a new learning task. This also influences the manner in which the human player explores a virtual environment, especially when it contains elements resembling real-world objects (e.g. ladders, monsters, keys, doors, etc.) For example, when the game avatar is located on a ladder, a human player will only use the up and down actions without having to learn to do so by exploring the game states. The player already knows, from real life, that other actions are not useful, as they do not lead to other states in the game.

\subsection{Private Eye}
The second game chosen for our evaluation is Private Eye. It is a game that features an avatar for a private investigator who is driving a car that can move around and jump vertically or over obstacles. The game environment can be seen as a labyrinth as well, as it consists of multiple roads located in a city or in the woods. In the city there are buildings near the roads, some of which play a special role in the game. The objective of the game is to capture thieves that sit behind the windows of different buildings. The thieves appear briefly and the player must move and "touch" the area near a thief in order to capture him/her. Capturing thieves provides the player with rewards and sometimes with special items, that must be returned to specific buildings (e.g. bags of money need to be returned to the bank, guns returned to the gun store). The player starts the game with 1000 points, but is penalized if he bumps into obstacles like birds and mice, or is attacked by thieves. The concept of multiple lives is not present in this game, however there is a three minute time cap for solving a game level; when the time limit has been reached, the game ends.

The game features similar exploration difficulties present in Montezuma's Revenge. As a consequence, this game has also proven to be a challenge for current deep reinforcement learning methods. The presence of a memory component for optimal play is even more important, as the player must travel long distances between collecting game items and dropping them at the appropriate locations. Some portions of the game look identical, and there is also a certain order in which the tasks should be carried out. For example, the game features a thief that must be captured and brought to the police building, but this can only be done after all the items in the current level have been returned where they belong. Capturing this final thief provides the highest in-game reward.

\subsection{Discussion} \label{difficulties}
The more challenging games from the Atari console (such as Montezuma's Revenge and Private Eye) present a particular difficulty compared to the simpler ones: the agent is not penalized for standing still. This could be yet another factor that further prevents efficient exploration.  In contrast, for games like Breakout or Pong (Figure \ref{procstructfig}), repeatedly choosing the no-op action will quickly lead to the end of the game or to losing points. The agent can thus learn to avoid this action, as it will be associated with a low utility value.

\section{DEEP REINFORCEMENT LEARNING WITH HUMAN CHECKPOINT REPLAY}

In this section we provide a description of the methods that we propose for training deep reinforcement learning agents on the Atari domain, as well as the architecture of the convolutional network that we used for the training process.

\subsection{Deep Reinforcement Learning}
Deep reinforcement learning \cite{mnih2013playing, mnih2015human} was the first method able to learn successful control policies directly from high-dimensional visual input on the Atari domain. It consists of a convolutional neural network that extracts features from the game frames and approximates the following action-value function

\begin{multline}
\label{eq:state-action-value}
Q^{*}(a, b) = \\ \max_{\pi}\E[r_{t} + \gamma r_{t+1} + \gamma^{2} r_{t+2} + ... \mid s_{t} = s, a_{t} = a, \pi]
\end{multline}

The computed value represents the sum of rewards $r_t$ discounted by $\gamma$ at each time step $t$, using a policy $\pi$ for the observation $s$ and action $a$. To solve the instability issue that reinforcement learning presents when a neural network is used to approximate the state-value function, experience replay \cite{lin1993reinforcement} is used, as well as a target network \cite{mnih2015human}. In order to train the network, Q-learning updates are applied on minibatches of experience, drawn at random from the replay memory. The Q-learning update at iteration $i$ uses the following loss function

\begin{multline}
\label{eq:loss}
L_{i}(\theta_{i}) = \\ \E_{(s, a, r, s') \sim U(D)} [(r + \gamma \max_{a'} Q(s', a'; \theta^{-}_{i}) - Q(s, a; \theta_{i}))^{2}]
\end{multline}

Here $\gamma$ is the discount factor determining the agent’s horizon, $\theta_i$ are the parameters of the Q-network at iteration $i$ and $\theta_i^-$ are the network parameters used to compute the target at iteration $i$. Differentiating the loss function with respect to the network weights gives the following gradient:

\begin{multline}
\label{eq:gradient}
\nabla_{\theta_{i}} L(\theta_{i}) =  \E_{s, a, r, s'} [(r + \gamma \max_{a'} Q(s', a'; \theta^{-}_{i}) - \\ - Q(s, a; \theta_{i}) \nabla_{\theta_{i}} Q(s, a; \theta_{i})].
\end{multline}

Our proposed approaches use the deep reinforcement learning architecture from \cite{mnih2015human}. Almost all the hyperparamaters have the same values, with the exception of the final exploration frame, for which we used a value of $4,000,000$ instead of $1,000,000$. We empirically found that using this value results in a slightly better performance of the agents. This may be caused by the fact that the most difficult games are also characterized by greater complexity and size of the state space, therefore a slower annealing of $\varepsilon$ may be helpful. Due to hardware limitations and the amount of time required for training deep Q-networks, we did not test for other values of this hyperparameter. Another structural difference we need to mention is the fact that, for the \textit{human experience replay} method presented next, we employed an additional replay memory.

\subsection{Human Checkpoint Replay}
For the most difficult games from the Atari platform that are characterized by sparse rewards, the original deep reinforcement learning approach \cite{mnih2013playing, mnih2015human} is not able to achieve positive scores. Thus the agents trained using deep reinforcement learning perform no better than a random agent.
% TR: este bine sa ne legam de ce am scris anterior si de asta am reformulat mai jos
As seen in the previous section, these games start in a state from which reaching the first reward is a long and challenging process for any player which does not possess any prior knowledge (such as commonsense world knowledge). Also considering the 18 actions available to the agent in each state, the $\varepsilon$-greedy strategy fails to find any game paths to a first state with positive reward. This hinders the convolutional neural network to learn relevant features to separate reward-winning states and actions from the bulk. Drawing inspiration from curriculum learning \cite{bengio2009curriculum} and the human starts evaluation metric used for testing Atari agents \cite{nair2015massively}, we introduce the \textit{human checkpoint replay} method. This consists of generating a number of checkpoints from human experience in the Arcade Learning Environment \cite{bellemare2012arcade} and storing them to be used as starting points for the training of deep reinforcement learning agents. Instead of resetting the environment to the beginning of the game each time a game episode ends, a checkpoint is randomly sampled from the checkpoint pool and restored in ALE.

The intuition behind this approach is that at least some of the checkpoints will have a reward located close enough in the game space for the $\varepsilon$-greedy strategy to be able to reach it. This way, the convolutional neural network is able to learn relevant features from the game frames and then successful control policies. As the training process advances, these will help the agent to become gradually more capable to reach rewards that are located farther away from the start state. Our method can also be thought of as being related to the planning concepts of landmarks \cite{karpas2009cost} and probabilistic roadmaps \cite{agha2013firm}.

\subsection{Human Experience Replay}
As a possible solution to the inefficiency of $\varepsilon$-greedy exploration in sparse reward environments, we also proposed training a deep reinforcement learning agent using offline human experience, combined with online agent experience. We dubbed this approach \textit{human experience replay}. It consists of storing human gameplay experience in same form of $(s,a,r,s')$ tuples in a separate replay memory and using it along with the original replay memory containing agent experience. This is meant to provide the agent with training samples that result in a positive reward, therefore making learning possible in environments that feature a sparse reward signal. The training process consists of repeatedly sampling a minibatch composed of both human transitions and agent transitions.

\section{EXPERIMENTS}
This section provides a thorough description of the experiments performed using our proposed approach for the two selected Atari games, Montezuma's Revenge and Private Eye. As mentioned earlier, we have chosen these two games because they are among the most challenging games on the Atari platform, even for human players. At this point no computer strategy has been able to learn an exploration technique which is better than a random agent, mainly due to the fact that no strategy is able to learn a solution which is able to reach any reward in the game. Therefore, any progress made towards solving these games will provides useful insights in the quest of developing general purpose agents.

\begin{figure}
\centerline{\includegraphics[height=1.35in]{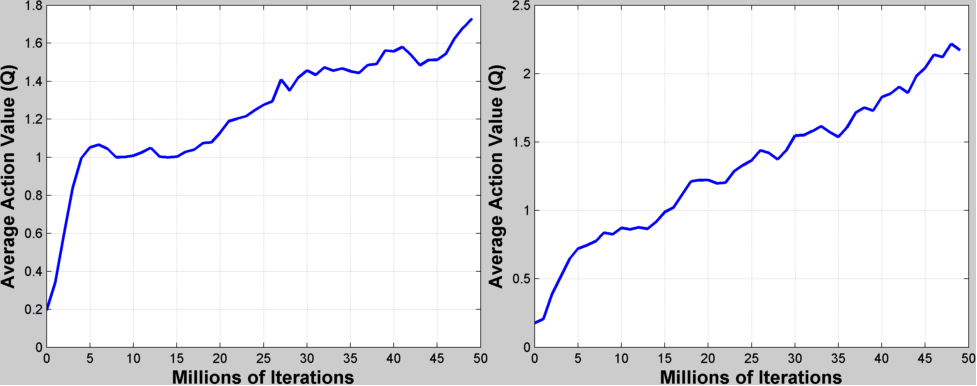}}
\caption{The two plots show the average maximum predicted action-value during training for our HCR DQN method on Montezuma's Revenge \textit{(left)} and Private Eye \textit{(right)}} 
\label{avgq}
\end{figure}

\subsection{Human Checkpoint Replay} \label{hcr}
The Arcade Learning Environment provides the capability of generating checkpoints during gameplay. These make it possible to continue running an environment from a given state at a later time by restoring a specific checkpoint in the emulator. The checkpoint consists of the memory content of the Atari 2600 console.

For the human checkpoint replay method (HCR DQN), we generated 100 checkpoints from a human player's experience for each game, stored them in an external file and then used them as starting points for the environment at training time, as well as for testing. The checkpoints that we used for training and testing, as well as the code for training deep reinforcement learning agents with human checkpoint replay, are publicly available \footnote{Code and checkpoints are available here: https://github.com/ionelhosu/atari-human-checkpoint-replay}. 
% TR: Daca avem spatiu, poate ar trebui sa spunem ca DeepMind sau alte lucrari nu au facut public codul respectiv.
We trained our networks using the generated checkpoints and performing Q-learning updates as describe in \cite{mnih2015human} for 50 million frames on each game. The two plots in Figure \ref{avgq} show how the average predicted $Q$ evolves during training on the games Montezuma's Revenge and Private Eye.

As discussed in section \ref{difficulties}, in difficult games such as Montezuma's Revenge and Private Eye, the avatar is not penalized for repeatedly choosing the no-op action. This raises two major issues. First of all, efficient exploration is prevented due to the neutral effect of repeatedly taking the no-op action. Also, in a deep reinforcement learning setting in which experience replay is used, by repeatedly choosing the no-op action, the replay memory will consistently be filled with transitions that are not relevant for the learning process. In order to avoid this outcome, we limit each training episode to 1800 frames, corresponding to 30 seconds of gameplay. By doing this, we make sure that the replay memory is populated with transitions that are relevant for the training process, as the agent will eventually be placed in checkpoints from which rewards are more easily accessible.

\subsection{Human Experience Replay}
Using ALE, we generated 1.2 million frames of human experience (about 5.5 hours of gameplay) for Montezuma's Revenge, consisting of $(s,a,r,s')$ transition tuples. Human experience transitions were stored in an additional replay memory during training. We performed Q-learning updates for 15 million frames on minibatches of size 32, composed of 16 samples of human experience and 16 samples of online agent experience. Due to the long training times required for training deep Q-networks and the cumbersome process of generating multiple hours of human experience, we only tested this approach on Montezuma's Revenge.

\subsection{Evaluation procedure}
In this paper, we used the human starts evaluation metric \cite{nair2015massively} to test the performance of the agents. The metric consists of using random checkpoints sampled from human experience as starting points for the evaluation of an agent. More specifically, we use a set of 100 checkpoints as human-generated start frames. In order to prove the robustness of our agent, the set of checkpoints used for evaluation are different than  the ones that were used for training. This evaluation method averages the score over 100 evaluations of 30 minutes of game time. The value of  $\varepsilon$ was fixed to $0.05$ throughout the evaluation process.

The random agent's scores were obtained using the same evaluation procedure. However the next action  to be performed in the environment was sampled from an uniform distribution.

\subsection{Quantitative Results}
As it can be observed in Table \ref{results}, the human checkpoint replay method provides a substantial improvement over a random agent for both games. In Montezuma's Revenge it obtains more than double the points of a random agent. In Private Eye, a random agent is not able to obtain a positive score, due to the multitude of negative rewards present in the game which the agent is not able to avoid. Our HCR DQN agent obtains significantly better results, demonstrating the success of this approach.

Compared to the HCR DQN agent, the human experience replay method provides only slightly better performance over a random agent in Montezuma's Revenge. Due to sparsity of rewards during the game, human experience alone cannot provide enough transitions that lead to positive rewards in order to facilitate learning, although it does provide a slightly better exploration compared to the random agent.

\begin{table} 
\begin{center}
\caption{Results obtained by our methods, human checkpoint replay (HCR DQN) and human experience replay (HER DQN) on Montezuma’s Revenge and Private Eye. The results represent raw game scores and were obtained using the human starts evaluation metric.} \label{results}
% Eu as lasa aici prescurtarile ca sa fie caption-ul mai scurt si as pune abrevierile si in text intr-una din sectiunle anterioare.
\begin{tabular}{lccc}
\hline
\rule{0pt}{10pt}
 &Random Agent &\multicolumn{1}{c}{\textbf{HCR DQN}}&\multicolumn{1}{c}{HER DQN}
\\
\hline
\\[-6pt]
Montezuma's Revenge &177.1&\textbf{379.1}& 218 \\
\hline
\\[-6pt]
Private Eye &--41&\textbf{1264.4}& N/A \\
\hline
\end{tabular}
\end{center}
\end{table}

\subsection{Qualitative Results}
We can draw better insights in the exploration of the agents by taking a closer look on the actions chosen by an agent. For Montezuma's Revenge, the HCR DQN agent is successfully collecting nearby rewards for all start points. For example, in the initial room of level 1, it successfully learns to climb the leftmost ladder in order to get the key. However, the agent still does not learn to avoid monsters and objects that lead to hypothetical negative rewards (such as losing a game life). This is mostly due to the fact that the game does not feature any negative rewards seen as changes in the score. This makes it even more difficult to find a successful exploration policy. While ALE offers this possibility, we did not provide our agents with an additional reward signal which penalizes the agent when it loses a life. It is also important to mention that in the majority of ALE checkpoints generated for training on Montezuma's Revenge there is no reward nearby. As a consequence, this set of checkpoints will continue to provide a challenge for future architectures, as it preserves much of the game's initial difficulty.

In Private Eye, the HCR DQN agent successfully collects most of the nearby rewards, and is seen successfully avoiding objects that lead to negative rewards. The agent does a great job at the latter task, especially as some of the negative rewards the avatar must avoid are moving fast and in an unpredictable manner, making this a difficult task even for human players.
% TR: cred ca si aici ar trebui spus ce nu invata ok

\subsection{Discussion}
Using human checkpoint replay might be seen as a trade-off for developing general game playing agents. The main objection would be that the agent uses human-generated start points for training the exploration model, which can be seen by some as an "deus ex-machina" intervention for the agent. However, the checkpoint replay merely provides additional starting points and does not offer an understanding of the explored game. It uses the experience of a human player to reach "easier" starting points, but for more difficult games this kind of intervention might be needed.

We should take into consideration that human players also make use of commonsense knowledge for these more difficult games. Using checkpoint replay does not bring any of this prior knowledge directly to the agent, maybe only in an indirect fashion as the human used it to get to that specific game positions. 

% TR aici trebuie sa spunem ca checkpoint-urile pot oferi echivalentul unui inceput de commonsense knowledge posedat de oameni si care ajuta la explorare. Nu este chiar perfect adevarat, dar are oarecare sens.

\section{CONCLUSION}
In this paper we presented a novel method using deep reinforcement learning, called human checkpoint replay, which was designed for some of the most difficult Atari 2600 games from the Arcade Learning Environment. Our experiments show a substantial improvement compared to all previous learning approaches, as well as over a random player. Our method draws inspiration from curriculum learning and it serves the purpose of compensating for the difficulties of current exploration strategies to find successful control policies in environments with sparse rewards.

As the results show, this method is a promising path of research. We will continue to study other approaches that deal with incentivizing and facilitating exploration in the most difficult games from the Atari platform. We believe that successfully learning control policies in such environments is closely related to the problem of achieving artificial general intelligence as in most real-life situations rewards are not encountered very frequently.

\bibliography{ecai}
\end{document}